\documentclass[10pt,twocolumn,letterpaper]{article}

\usepackage{iccv}
\usepackage{times}
\usepackage{epsfig}
\usepackage{graphicx}
\usepackage{amsmath}
\usepackage{amssymb}
\usepackage{algorithm}
\usepackage{algorithmic}

\usepackage[breaklinks=true,bookmarks=false]{hyperref}

\iccvfinalcopy 

\def\iccvPaperID{****} 

\ificcvfinal\fi

\begin{document}


\title{Deep Self-Learning From Noisy Labels}
\vspace{-3mm}
\author{Jiangfan Han$^{1}$ \quad Ping Luo$^{2}$ \quad Xiaogang Wang$^{1}$ \\
$^{1}$CUHK-SenseTime Joint Laboratory, The Chinese University of Hong Kong \\ $^{2}$The University of Hong Kong\\
{\tt\small \{jiangfanhan@link., xgwang@ee.\}cuhk.edu.hk \quad\quad pluo@cs.hku.hk}
}

\maketitle
\ificcvfinal\thispagestyle{empty}\fi

\begin{abstract}
   ConvNets achieve good results when training from clean data, but learning from noisy labels significantly degrades performances and remains challenging. Unlike previous works constrained by many conditions, making them infeasible to real noisy cases, this work presents a novel deep self-learning framework to train a robust network on the real noisy datasets without extra supervision. The proposed approach has several appealing benefits. (1) Different from most existing work, it does not rely on any assumption on the distribution of the noisy labels, making it robust to real noises. (2) It does not need extra clean supervision or accessorial network to help training. (3) A self-learning framework is proposed to train the network in an iterative end-to-end manner, which is effective and efficient. Extensive experiments in challenging benchmarks such as Clothing1M and Food101-N show that our approach outperforms its counterparts in all empirical settings.
\end{abstract}

\vspace{-4mm}
\section{Introduction}

Deep Neural Networks (DNNs) achieve impressive results on many computer vision tasks such as
image recognition~\cite{krizhevsky2012imagenet,szegedy2015going,taigman2014deepface}, semantic segmentation~\cite{long2015fully,zhao2017pyramid,noh2015learning}, object detection~\cite{girshick2015fast,ren2015faster,redmon2017yolo9000,lin2017feature} and cross modality tasks~\cite{liu2018show,liu2019improving,zhou2019talking}.
However, many of these tasks require large-scale datasets with reliable and clean annotations to train DNNs such as ImageNet~\cite{deng2009imagenet} and MS-COCO~\cite{lin2014microsoft}.
But collecting large-scale datasets with precise annotations is expensive and time-consuming, preventing DNNs from being employed in real-world noisy scenarios.
%
Moreover, most of the ``ground truth annotations'' are from human labelers, who also make mistakes and increase biases of the data.
%
%

An alternative solution is to collect data from the Internet by using different image-level tags as queries. These tags
can be regarded as labels of the collected images.
This solution is cheaper and more time-efficient than human annotations, but the collected labels may contain noises.
A lot of previous work has shown that noisy labels lead to an obvious decrease in performance of DNNs~\cite{xiao2015learning,nettleton2010study,pechenizkiy2006class}.
Therefore, attentions have been concentrated on how to improve the robustness of DNNs against noisy labels.
%
\begin{figure}[t]
\begin{center}
\includegraphics[width=7.6cm ,scale=0.54]{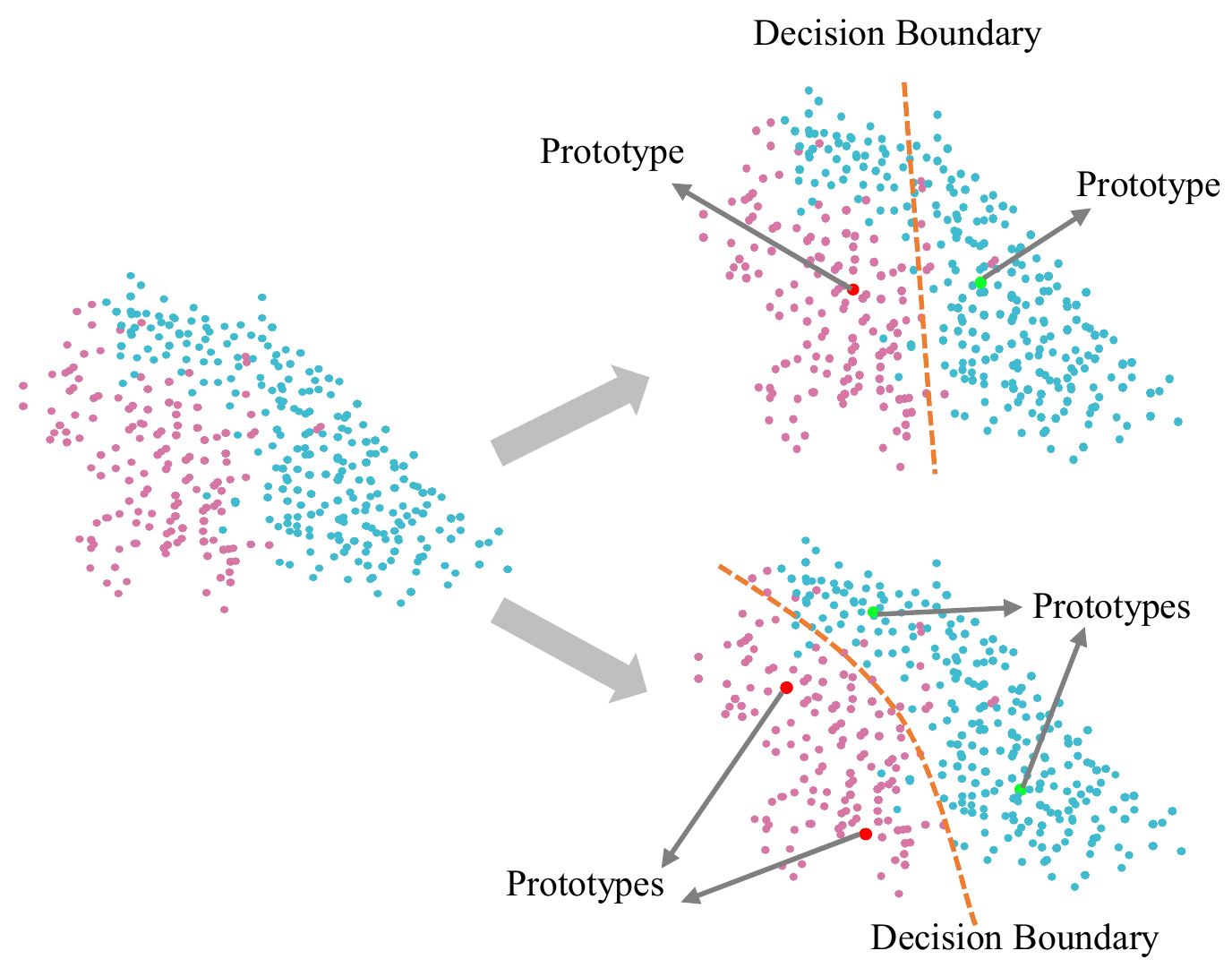}
\end{center}
\vspace{-3mm}
   \caption{An example of solving two classes classification problem using different number of prototypes. \textbf{Left}: Original data distribution. Data points with the same color belong to the same class. \textbf{Upper Right}: The decision boundary obtained by using a single prototype for each class. \textbf{Lower Right}: The decision boundary obtained by two prototypes for each class. Two prototypes for each class leads to a better decision boundary.}
\label{fig:proto_vis}
\vspace{-6mm}
\end{figure}

Previous approaches tried to correct the noisy labels by introducing a transition matrix~\cite{patrini2017making,hendrycks2018using} into their loss functions, or by adding additional layers to estimate the noises~\cite{goldberger2016training,sukhbaatar2014training}.
Most of these methods followed a simple assumption to simplify the problem: There is a single transition probability between the noisy label and ground-truth label, and this probability is independent of individual samples. 
But in real cases, the appearance of each sample has much influence on whether it can be misclassified.
Due to this assumption, although these methods worked well on hand-crafted noisy datasets such as CIFAR10~\cite{krizhevsky2009learning} with manually flipped noisy labels,
their performances were limited on real noisy datasets such as Clothing1M~\cite{xiao2015learning} and Food101-N \cite{lee2018cleannet}.

Also, noisy tolerance loss functions~\cite{tanaka2018joint,zhang2018generalized} have been developed to fight against label noises, but they had a similar assumption as the above noise correction approaches. 
So they were also infeasible for real-world noisy datasets.
Furthermore, many approaches~\cite{lee2018cleannet,li2017learning,veit2017learning} solved this problem by using additional supervision.
For instance, some of them manually selected a part of samples and asked human labelers to clean these noisy labels.
By using extra supervision, these methods could improve the robustness of deep networks against noises.
The main drawback of these approaches was that they required extra clean samples, making them expensive to apply in large-scale real-world scenarios.

Among all the above work, CleanNet~\cite{lee2018cleannet} achieved the existing state-of-the-art performance on real-world dataset such as Clothing1M \cite{xiao2015learning}.
CleanNet used ``class prototype'' (\ie a representative sample) to represent each class category and decided whether the label for a sample is correct or not by comparing with the prototype.
However, CleanNet also needed additional information or supervision to train.

To address the above issues, we propose a novel framework of Self-Learning with Multi-Prototypes (SMP), which aims to train a robust network on the real noisy dataset without extra supervision.
By observing the characteristics of samples in the same noisy category, we conjecture that these samples have widely spread distribution.
A single class prototype is hard to represent all characteristics of a category. 
More prototypes should be used to get a better representation of characteristics.
Figure~\ref{fig:proto_vis} illustrated the case and further exploration has been conducted in the experiment.
Furthermore, extra information (supervision) is not necessarily available in practice.

The proposed SMP trains in an iterative manner which contains two phases: the first phase is to train a network with the original noisy label and corrected label generated in the second phase. 
The second phase uses the network trained in the first stage to select several prototypes. These prototypes are used to generate the corrected label for the first stage.
This framework does not rely on any assumption on the distribution of noises, which makes it feasible to real-world noises. It also does not use accessorial neural networks nor require additional supervision, providing an effective and efficient training scheme. 

The contributions of this work are summarized as follows.
(1) We propose an iterative learning framework SMP to relabel the noisy samples and train ConvNet on the real noisy dataset, without using extra clean supervision. Both the relabeling and training phases contain only one single ConvNet that can be shared across different stages, making SMP effective and efficient to train.
(2) SMP results in interesting findings for learning from noisy data. For example, unlike previous work \cite{lee2018cleannet}, we show that a single prototype may not be sufficient to represent a noisy class.
By extracting multiple prototypes for a category, we demonstrate that more prototypes would get a better representation of a class and obtain better label-correction results.
(3) Extensive experiments validate the effectiveness of SMP on different real-world noisy datasets.
We demonstrate new state-of-the-art performance on all these datasets. 

\vspace{-1mm}
\section{Related Work}
\vspace{-1mm}
\textbf{Learning on noisy data.}
ConvNets achieved great successes when training with clean data. However, the performances of ConvNets degraded inevitably when training on the data with noisy labels~\cite{nettleton2010study,pechenizkiy2006class}.
The annotations provided by human labelers on websites such as Amazon Mechanical Turk~\cite{ipeirotis2010quality} would also introduce biases and incorrect labels.
As annotating large-scale clean and unbias dataset is expensive and time-consuming,
many efforts have been made to improve the robustness of ConvNets trained on noisy datasets.
They can be generally summarized as three parts mentioned below.

First, the transition matrix was widely used to capture the transition probability between the noisy label and true label,
\ie the sample with a true label $y$ has a certain probability to be mislabeled as a noisy label $\tilde y$.
Sukhbaatar \etal in~\cite{sukhbaatar2014training} added an extra linear layer to model the transition relationships between true and corrupted labels.
Patrini \etal in~\cite{patrini2017making} provided a loss correction method to estimate the transition matrix by using a deep network trained on the noisy dataset.
The transition matrix was estimated by using a subset of cleanly labeled data in~\cite{hendrycks2018using}.
The above methods followed an assumption that the transition probability is identical between classes and is irrelevant to individual images.
Therefore, these methods worked well on the noisy dataset that is created intentionally by human with label flipping such as the noisy version of CIFAR10~\cite{krizhevsky2009learning}.
However, when applying these approaches to real-world datasets such as Clothing1M~\cite{xiao2015learning}, their performances were limited since the assumption above is no longer valid.

Second, another scenario was to explore the robust loss function against label noises. \cite{ghosh2015making} explored the tolerance of different loss functions under uniform label noises.
Zhang and Sabuncu~\cite{zhang2018generalized} found that the mean absolute loss function is more robust than the cross-entropy loss function, but it has other drawbacks.
Then they proposed a new loss function that benefits both of them.
However, these robust loss functions had certain constraints so that they did not perform well on real-world noisy datasets.

Third, CleanNet~\cite{lee2018cleannet} designed an additional network to decide whether a label is noisy or not.
The weight of each sample during network training is produced by the CleanNet to reduce the influence of noisy labels in optimization.
Ren \etal~\cite{ren2018learning} and Li \etal~\cite{li2019learning} tried to solve noisy label training by meta-learning.
Some methods~\cite{guo2018curriculumnet,jiang2018mentornet} based on curriculum learning were also developed to train against label noises.
CNN-CRF model was proposed by Vahdat~\cite{vahdat2017toward} to represent the relationship between noisy and clean labels.
However, most of these approaches either required extra clean samples as additional information or adopted a complicated training procedure.
In contrast, SMP not only corrects noisy labels without using additional clean supervision but also trains the network in an efficient end-to-end manner, achieving state-of-the-art performances on both Clothing1M \cite{xiao2015learning} and Food101-N \cite{lee2018cleannet} benchmarks.
When equipped with a few additional information, SMP further boosts the accuracies on these datasets.

\textbf{Self-learning by pseudo-labels.}
Pseudo-labeling~\cite{ding2018semi,tanaka2018joint,lee2013pseudo} belongs to the self-learning scenario, and 
it is often used in semi-supervised learning where the dataset has a few labeled data and most of the data are unlabeled.
In this case, the pseudo-labels are given to the unlabeled data by using the predictions from the model pretrained on labeled data.
In contrast, when learning from noisy datasets, all data have labels, 
but they may be incorrect.
%
Reed \etal \cite{reed2014training} proposed to jointly train noisy and pseudo-labels.
However, the method proposed in \cite{reed2014training} over-simplifies the assumption of the noisy distribution, 
leading to a sub-optimal result.
Joint Optimization~\cite{tanaka2018joint} completely replaced all labels by using pseudo-labels.
However, \cite{tanaka2018joint} discarded the useful information in the original noisy labels.
In this work, we predict the pseudo-labels by using SMP and train deep network by using both the original labels and pseudo-labels in a self-learning scheme.
\begin{figure*}[t]
\begin{center}{}
\includegraphics[width= 17.9cm ,scale=0.56]{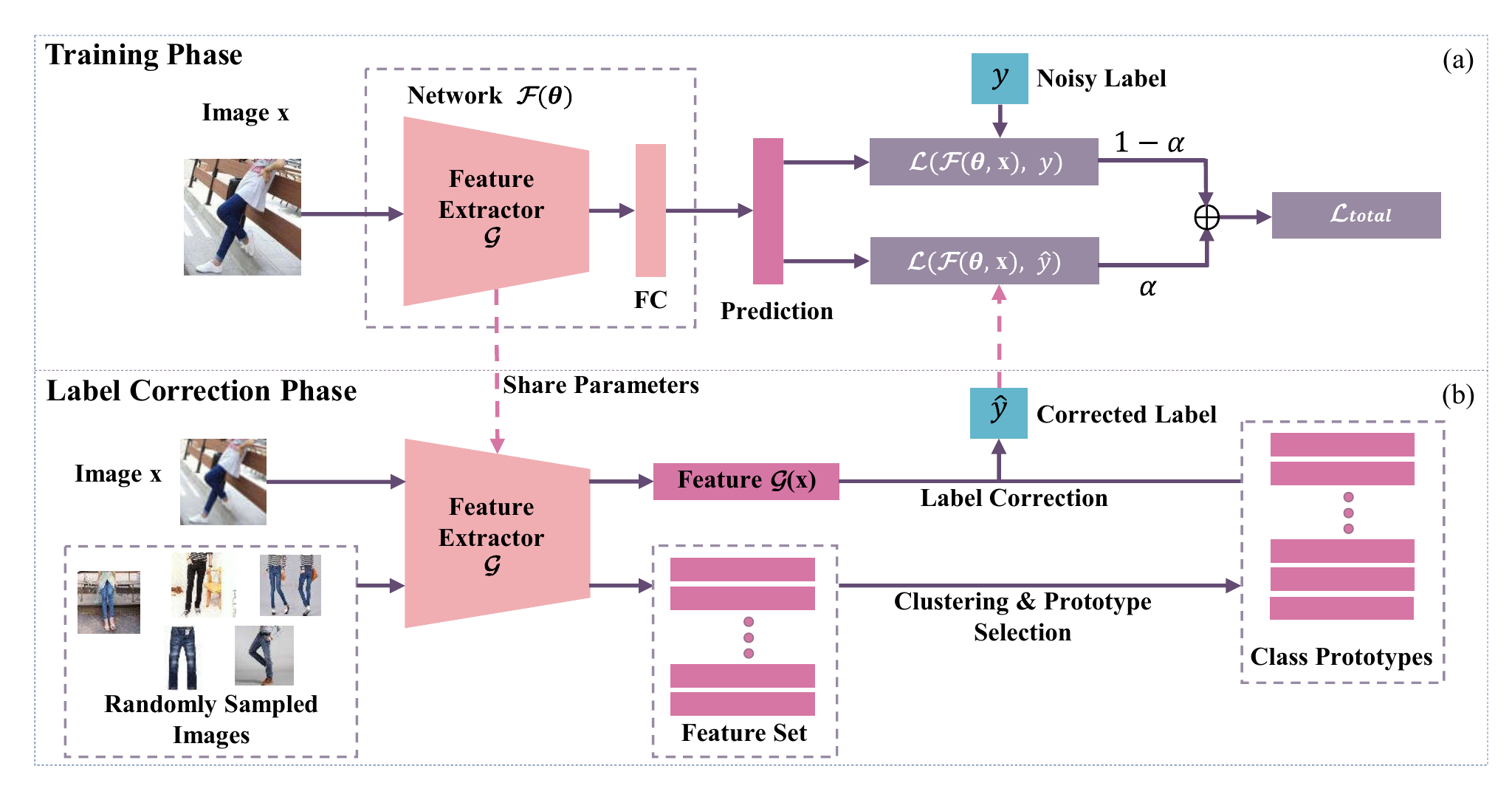}
\end{center}
\vspace{-5mm}
   \caption{Illustration of the pipeline of iterative self-learning framework on the noisy dataset. (a) shows the training phase and (b) shows the label correction phase, where these two phases proceed iteratively. The deep network $\mathcal{G}$ can be shared, such that only a single model needs to be evaluated in testing.}
\label{fig:overview}
\vspace{-4mm}
\end{figure*}
\vspace{-1mm}
\section{Our Approach}
\vspace{-1mm}

\textbf{Overview.} Let $\mathcal{D}$ be a noisily-labeled dataset, $\mathcal{D} = \{\mathbf{X}, Y\} = \{(\mathbf{x}_{1}, y_1),...,(\mathbf{x}_{N}, y_N)\}$, which contains $N$ samples, and $y_{i} \in \{1,2,..., K\}$ is the noisy label corresponding to the image $\mathbf{x}_{i}$. $K$ is the number of classes in the dataset.
Since the labels are noisy, they would be incorrect, impeding model training.
To this end, a neural network $\mathcal{F}(\theta)$ with parameter $\theta$ is defined to transform the image $\mathbf{x}$ to the label probability distribution $\mathcal{F}(\theta, \mathbf{x})$.
When training on a cleanly-labeled dataset, an optimization problem is defined as
\begin{align}
   \theta^{*} = \mathrm{argmin}_{\theta} \ \mathcal{L}(Y, \mathcal{F}(\theta, \mathbf{X}))
\label{eq:obj}
\end{align}
where $\mathcal{L}$ represents the empirical risk.
However, when $Y$ contains noises, the solution of the above equation would be sub-optimal.
When label noises are presented, all previous work that improved the model robustness can be treated as adjusting the term in Eqn.\eqref{eq:obj}.
In this work, we propose to attain the corrected label $\hat{Y}(\mathbf{X}, \mathbf{X}_s)$ in a self-training manner,
where $\mathbf{X}_s$ indicates a set of class prototypes to represent the distribution of classes.
Our optimization objective is formulated as
\begin{align}
   \theta^{*} = \mathrm{argmin}_{\theta} \ \mathcal{L}(Y, \hat{Y}(\mathbf{X}, \mathbf{X}_s), \mathcal{F}(\theta, \mathbf{X}))
\label{eq:obj_our}
\end{align}
Although the corrected label $\hat{Y}(\mathbf{X},\mathbf{X}_s)$ is more precise than the original label $Y$,
we believe that it is still likely to misclassify the hard samples as noises. So we keep the original noisy label $Y$ as a part of supervision in the above objective function.

The corrected label $\hat{y}_i(\mathbf{x}_i, \mathbf{X}_s)\in \hat{Y}(\mathbf{X},\mathbf{X}_s)$ of image $\mathbf{x}_i$ is given by a similarity metric between the image $\mathbf{x}_i$ and the set of prototypes $ \mathbf{X}_s$.
Since the data distribution of each category is complicated, a single prototype is hard to represent the distribution of the entire class.
We claim that using multi-prototypes can get a better representation of the distribution, leading to better label correction.

In the following sections, we introduce the iterative self-learning framework in details, where a deep network learns from the original noisy dataset,
and then it is trained to correct the noisy labels of images.
The corrected labels will supervise the training process iteratively.

\subsection{Iterative Self-Learning}

\textbf{Pipeline}. The overall framework is illustrated in Figure~\ref{fig:overview}.
It contains two phases, the training phase, and the label-correction phase.
In the training phase, a neural network $\mathcal{F}$ with parameters $\theta$ is trained, taking image $\mathbf{x}$ as input and producing the corresponding label prediction $\mathcal{F}(\theta, \mathbf{x})$.
The supervision signal is composed by two branches, (1) the original noisy label $y$ corresponding to the image $\mathbf{x}$ and (2) the corrected label $\hat{y}$ generated by the second phase of label correction.

In the label correction phase, we extract the deep features of the images in the training set by using the network $\mathcal{G}$ trained in the first stage. Then we explore a selection scheme to select several class prototypes for each class. 
Afterward, we correct the label for each sample according to the similarity of the deep features of the prototypes.
The corrected labels are then used as a part of supervision in the first training phase. The first and the second phases proceed iteratively until the training converged.

\subsection{Training Phase}

The pipeline of the training phase is illustrated in Figure~\ref{fig:overview}~(a).
This phase aims to optimize the parameters $\theta$ of the deep network $\mathcal{F}$.
In general, the objective function is the empirical risk of cross-entropy loss, which is formulated by
\begin{align}
\vspace{-1mm}
    \mathcal{L}(\mathcal{F}(\theta, \mathbf{x}), y) = -\frac{1}{n} \sum_{i=1}^{n}\log(\mathcal{F}(\theta, \mathbf{x}_{i})_{y_i})
\label{ce-loss}
\vspace{-1mm}
\end{align}
where $n$ is the mini-batch size and $y_i$ is the label corresponding to the image $\mathbf{x}_i$.
When learning on a noisy dataset, the original label $y_i$ may be incorrect, so we introduce another corrected label as a complementary supervision.
The corrected label is produced by a self-training scheme in the label correction phase.
With the corrected signal, the objective loss function is
\begin{align}
    \mathcal{L}_{total} = (1-\alpha) \mathcal{L}(\mathcal{F}(\theta, \mathbf{x}), y) + \alpha \mathcal{L}(\mathcal{F}(\theta, \mathbf{x}), \hat{y})
\label{joint-loss}
\end{align}
where $\mathcal{L}$ is the cross entropy loss as shown in Eqn.(\ref{ce-loss}), $y$ is the original noisy label, and $\hat{y}$ is the corrected label produced by the second phase. The weight factor $\alpha \in [0,1]$ controls the important weight of the two terms.

Since the proposed approach does not require extra information (typically produced by using another deep network or additional clean supervision),
at the very beginning of training,
we set $\alpha$ to 0 and train the network $\mathcal{F}$ by using only the original noisy label $y$.
After a preliminary network was trained, we can step into the second phase and obtain the corrected label $\hat{y}$.
At this time, $\alpha$ is a positive value, where the network is trained jointly by $y$ and $\hat{y}$ with the objective shown in Eqn. (\ref{joint-loss}).

\subsection{Label Correction Phase}

In the label correction phase, we aim to obtain a corrected label for each image in the training set.
These corrected labels will be used to guide the training procedure for the first phase in turn.

For label correction, the first step is to select several class prototypes for each category.
Inspired by the clustering method~\cite{rodriguez2014clustering}, we propose the following method to pick up these prototypes.
(1) We use the preliminary network trained in the first phase to extract deep features of images in the training set.
In experiments, we employ the ResNet~\cite{he2016deep} architecture, where the output before the fully-connected layer is regarded as the deep features, denoted as $\mathcal{G}(\mathbf{x})$.
Therefore, the relationship between $\mathcal{F}(\theta, \mathbf{x})$ and $\mathcal{G}(\mathbf{x})$ is $\mathcal{F}(\theta, \mathbf{x}) = f(\mathcal{G}(\mathbf{x}))$, where $f$ is the operation on the fully-connected layer of ResNet.
%
(2) In order to select the class prototypes for the $c$-th class, we extract a set of deep features, $\{\mathcal{G}(\mathbf{x}_i)\}_{i=1}^n$, corresponding to a set of images $\{\mathbf{x}_i\}_{i=1}^n$ in the dataset with the same noisy label $c$.
Then, we calculate the cosine similarity between the deep features and construct a similarity matrix $\mathbf{S} \in \mathbb{R}^{n \times n}$, $n$ is the number of images with noisy label $c$\ and $S_{ij}\in\mathbf{S}$ with
\begin{align}
   S_{ij} = \frac{\mathcal{G}(\mathbf{x}_{i})^{T} \mathcal{G}(\mathbf{x}_{j})}{||\mathcal{G}(\mathbf{x}_{i})||_{2}||\mathcal{G}(\mathbf{x}_{j})||_{2}}
\end{align}
Here $S_{ij}$ is a measurement of the similarity between two images $\mathbf{x}_i$ and $\mathbf{x}_j$.
Larger $S_{ij}$ indicates the two images with higher similarity.
Both \cite{rodriguez2014clustering} and \cite{guo2018curriculumnet} used Euclidean distance as the similarity measurement, but we find that cosine similarity is a better choice to correct the labels.
The comparisons between the Euclidean distance and the cosine similarity are provided in experiment.
An issue is that the number of images $n$ in a single category is huge \eg $n=$70k for Clothing1M, making the calculation of this cosine similarity matrix $\mathbf{S}$ time-consuming. Furthermore, latter calculation using such a huge matrix is also expensive.
So we just randomly sample $m$ images ($m<n$) in the same class to calculate the similarity matrix $\mathbf{S}^{m\times m}$ to reduce the computational cost.
To select prototypes, we define a density $\rho_i$ for each image $\mathbf{x}_i$,
\begin{align}
   \rho_i = \sum_{j=1}^{m}\mathrm{sign}(S_{ij} - S_c)
\end{align}
where $\mathrm{sign}(x)$ is the sign function\footnote{We have $\mathrm{sign}(x)=1$ if $x>0$; $\mathrm{sign}(x)=0$ if $x=0$; otherwise $\mathrm{sign}(x)=-1$.}.
The value of $S_c$ is a constant number given by
the value of an element ranked top 40\% in $\mathbf{S}$ where the values of elements in $\mathbf{S}$ are ranked in an ascending order from small to large.
We find that the concrete choice of $S_c$ does not have influence in the final result, because we only need the relative density of images.

\begin{figure}[t]
    \centering
    \small
    \begin{tabular}{c@{\hspace{-6mm}}c@{\hspace{-3mm}}c}
        &
        \begin{tabular}{c}
            \includegraphics[width=4.5cm]{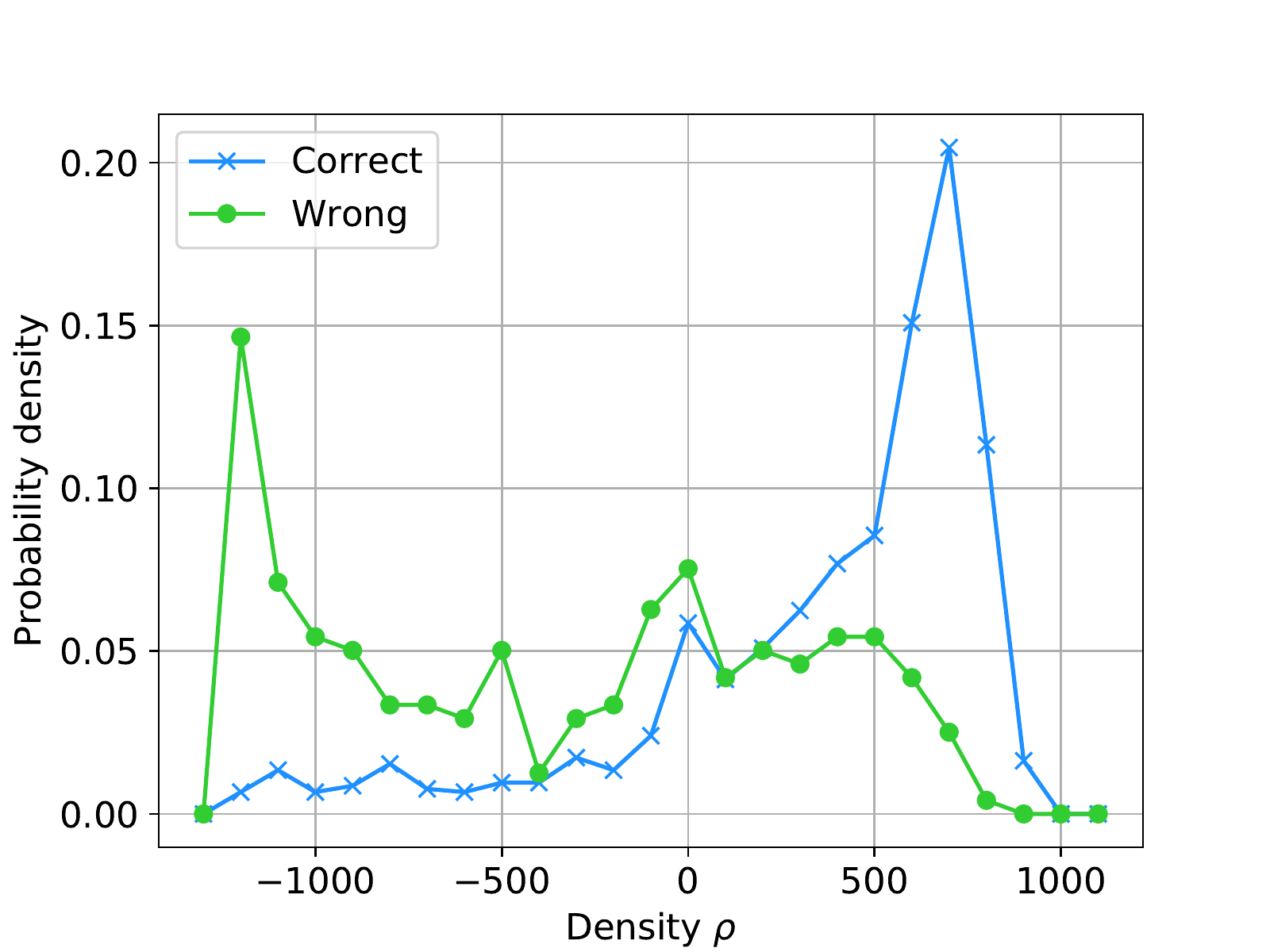}
        \end{tabular}
        &
        \begin{tabular}{c}
            \includegraphics[width=4.5cm]{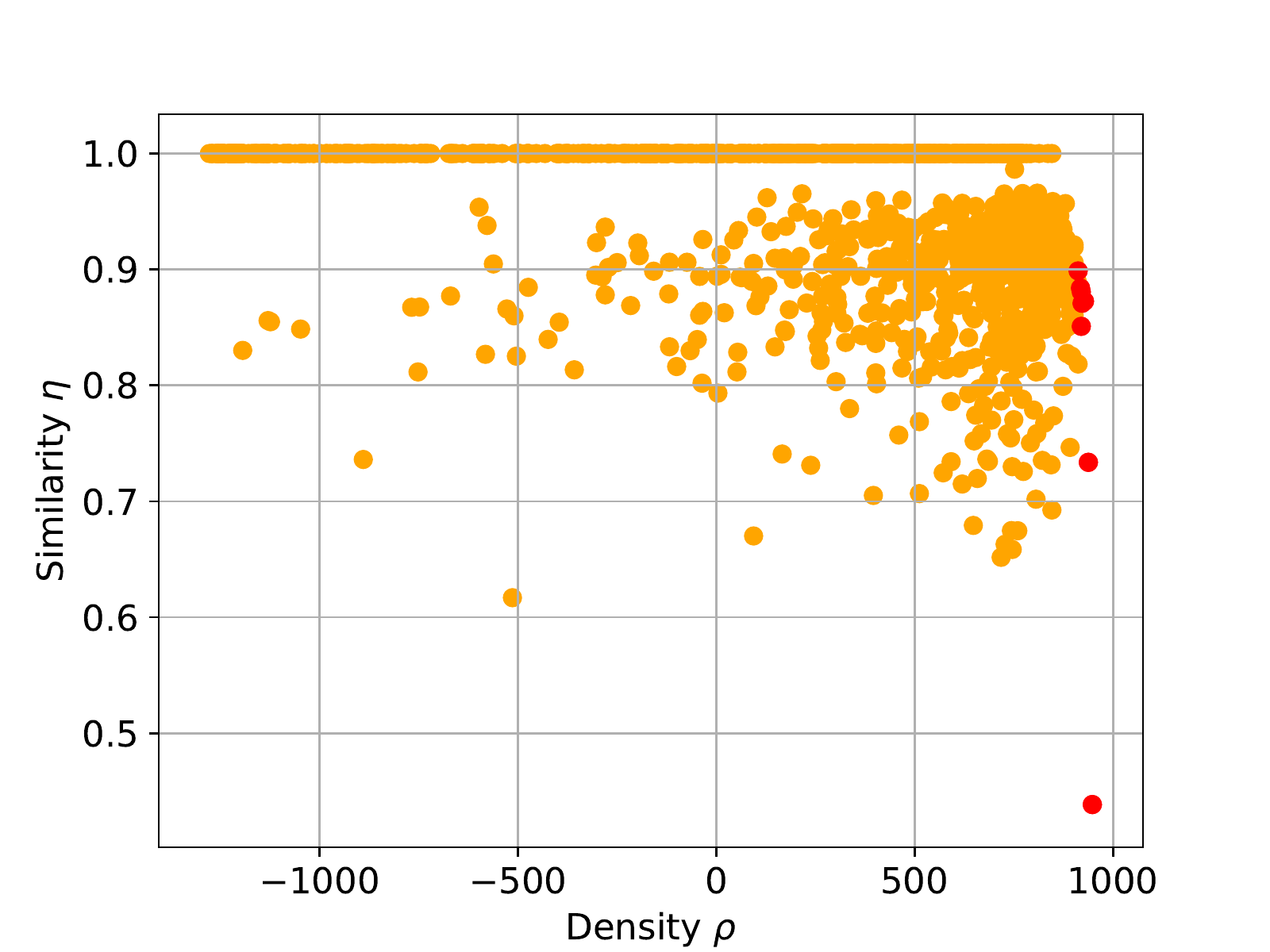}
        \end{tabular}\\
        &(a) & (b)\
    \end{tabular}
    \caption{(a)  The probability density with the density $\rho$ for sample with correct label (blue line) and sample with wrong label (green line) for 1280 images sampled from the same noisy class in Clothing1M dataset. (b) The distribution between similarity $\eta$ and density $\rho$. the samples are the same as (a). Red dots are samples with top-8 highest $\rho$ value.}
    \label{fig:density}
    \vspace{-4mm}
\end{figure}

\textbf{Discussions.} From the above definition of density $\rho$, the image with larger $\rho$ has more similar images around it.
These images with correct labels should be close to each other, while the images with noisy labels are usually isolated from others. The probability density with  $\rho$ for images with correct label and images with the wrong label is shown in Figure~\ref{fig:density}~(a).
We can find the images with correct labels are more possible to have large $\rho$ value while those images with wrong labels appear in the region with low $\rho$.   
In other words, the images with larger density $\rho$ have a higher probability to have the correct label in the noisy dataset and can be treated as prototypes to represent this class.
If we need $p$ prototypes for a class, we can regard the images with the top-$p$ highest density values as the class prototypes.

Nevertheless, the above strategy to choose prototypes has a weakness, that is,
if the chosen $p$ prototypes belonging to the same class are very close to each other, the representative ability of these $p$ prototypes is equivalent to using just a single prototype.
To avoid such case, we further define a similarity measurement $\eta_i$ for each image $\mathbf{x}_i$
\begin{align}
   \eta_i = \begin{cases}\max_{j,~\rho_j > \rho_i} S_{ij}, & \rho_i < \rho_{max} \cr \min_j S_{ij}, & \rho_i = \rho_{max}\end{cases}
\end{align}
where $\rho_{max} = \max\{\rho_1, ..., \rho_m\}$.
From the definition of $\eta$, we find that for the image $\mathbf{x}_i$ with density value equaled to $\rho_{max}$ ($\rho_i = \rho_{max}$), its similarity measure $\eta_i$ is the smallest.
Otherwise, for those images $\mathbf{x}_i$ with $\rho_i < \rho_{max}$, the similarity $\eta_i$ is defined as the maximum of the cosine similarity between the image $i$ with features $\mathcal{G}(\mathbf{x}_i)$ and the other image $j$ with features $\mathcal{G}(\mathbf{x}_j)$, whose density value is higher than $\mathbf{x}_i$ ($\rho_j > \rho_i$).

From the above definitions, smaller similarity value $\eta_i$ indicates that the features corresponding to the image $i$ are not too close the other images with density $\rho$ larger than it.
So, the sample with high-density value $\rho$ (probability a clean label), and low similarity value $\eta$ (a clean label but moderately far away from other clean labels) can fulfill our selection criterion as the class prototypes.
In experiments, we find that the samples with high density $\rho$ ranked the top often have relatively small similarity values $\eta$.

As shown in Figure~\ref{fig:density}~(b), red dots are samples with density $\rho$ ranked in the top. Over 80\% of the samples have $\eta>0.9$ and half of the samples have $\eta>0.95$. So those red dots have relative small $\eta$ value and far away from each other. 
It also proves our claim that the samples in the same class tend to gather in several clusters, so a single prototype is hard to represent an entire class and therefore more prototypes are necessary.
In experiments, we select the prototypes ranking in the top with $\eta<0.95$.

After the selection of prototypes for each class, we have a prototype set $\{\mathcal{G}(\mathbf{X}_1),...,\mathcal{G}(\mathbf{X}_c),...,\mathcal{G}(\mathbf{X}_K)\}$ (represented by deep features), where $\mathbf{X}_c = \{\mathbf{x}_{c1}, ..., \mathbf{x}_{cp}\}$ is the selected images for the $c$-th class, $p$ is the number of prototypes for each class, and $K$ is the number of classes in the dataset.
Given an image $\mathbf{x}$, we calculate the cosine similarity between extracted features $\mathcal{G}(\mathbf{x})$ and different sets of prototypes $\mathcal{G}(\mathbf{X}_c)$. The similarity score $\sigma_c$ for the $c$-th class is calculated as
\begin{align}
   \sigma_c = \frac{1}{p}{\sum_{l = 1}^p\cos(\mathcal{G}(\mathbf{x}), \mathcal{G}(\mathbf{x}_{cl}))},c=1...K
\end{align}
where $\mathcal{G}(\mathbf{x}_{cl})$ is the $l$-th prototype for the $c$-th class.
Here we use the average similarity over $p$ prototypes, instead of the maximum similarity, because we find that combination (voting) from all the prototypes might prevent misclassifying some hard samples with almost the same high similarity to different classes.
Then, we obtain the corrected label $\hat{y} \in \{1,\dots,K\}$ by
\begin{align}
   \hat{y} = \mathrm{argmax}_c \ \sigma_c,~~c=1...K
\end{align}
After getting the corrected label $\hat{y}$, we treat it as complementary supervision signal to train the neural network $\mathcal{F}$ in the training phase.
\begin{algorithm}
\caption{Iterative Learning}
\label{alg}
\begin{algorithmic}[1]
\STATE Initialize network parameter $\theta$
\FOR {$M = 1 : \mathrm{num\_epochs}$}
\IF {$M < \mathrm{start\_epoch}$}
\STATE sample ($\mathbf{X}, Y$) from training set.
\STATE $\theta^{(t+1)} \gets \theta^{(t)} -  \xi \nabla \mathcal{L}(\mathcal{F}(\theta^{(t)}, \mathbf{X}), Y)$
\ELSE
\STATE Sample \{$\mathbf{x}_{c1},\dots,\mathbf{x}_{cm}$\} for each class label $c$.
\STATE Extract the feature and calculate the similarity $\mathbf{S}$.
\STATE Calculate the density $\rho$ and elect the class prototypes $\mathcal{G}(\mathbf{X}_c)$ for each class $c$.
\STATE Get the corrected $\hat{y}$ for each sample $x_i$
\STATE sample ($\mathbf{X}, Y , \hat{Y}$) from training set.
\STATE $\theta^{(t+1)} \gets \theta^{(t)} -  \xi \nabla ((1-\alpha) \mathcal{L}(\mathcal{F}(\theta^{(t)}, \mathbf{X}), Y) + \alpha \mathcal{L}(\mathcal{F}(\theta^{(t)}, \mathbf{X}), \hat{Y})$
\ENDIF
\ENDFOR
\end{algorithmic}
\end{algorithm}

\subsection{Iterative Self-Learning}

As shown in Algorithm~\ref{alg}, the training phase and the label correction phase proceed iteratively.
The training phase first trains an initial network by using image $\mathbf{x}$ with noisy label $y$, as no corrected label $\hat{y}$ provided.
Then we proceed to the label correction phase. Feature extractor in this phase shares the same network parameters as the network $\mathcal{F}$ in the training phase.
We randomly sample $m$ images from the noisy dataset for each class and extract features by $\mathcal{F}$, and then the prototype selection procedure selects $p$ prototypes for each class.
Corrected label $\hat{y}$ is assigned to every image $\mathbf{x}$ by calculating the similarity between its features $\mathcal{G}(\mathbf{x})$ and the prototypes.
This corrected label $\hat{y}$ is then used to train the network $\mathcal{F}$ in the next epoch.
The above procedure proceeds iteratively until converged.
\vspace{-2mm}
\section{Experiments}
\vspace{-1mm}
\textbf{Datasets.} We employ two challenging real-world noisy datasets to evaluate our approach, Clothing1M~\cite{xiao2015learning} and Food101-N~\cite{lee2018cleannet}.
(1) \emph{Clothing1M}~\cite{xiao2015learning} contains 1 million images of clothes, which are classified into 14 categories.
The labels are generated by the surrounding text of the images on the Web, so they contain many noises.
The accuracy of the noisy label is 61.54\%.
Clothing1M is partitioned into training, validation and testing sets, containing 50k, 14k and 10k images respectively.
Human annotators are asked to clean a set of 25k labels as a clean set. In our approach, they are not required to use in training.
(2)
\emph{Food101-N~\cite{lee2018cleannet}} is a dataset to classify food. It contains 101 classes with 310k images searched from the Web.
The accuracy of the noisy label is 80\%.
It also provides 55k verification labels (clean by humans) in the training set. 

\textbf{Experimental Setup.}
For the Clothing1M dataset, we use ResNet50 pretrained on the ImageNet.
The data preprocessing procedure includes resizing the image with a short edge of 256 and randomly cropping a 224$\times$224 patch from the resized image.
We use the SGD optimizer with a momentum of 0.9. The weight decay factor is $5 \times 10^{-3}$, and the batchsize is 128.
The initial learning rate is 0.002 and decreased by 10 every 5 epochs. The total training processes contain 15 epochs.
In the label correction phase, we randomly sample 1280 images for each class in the noisy training set, and
8 class prototypes are picked out for each class.
For the Food-101N, the learning rate decreases by 10 every 10 epochs, and there are 30 epochs in total.
The other settings are the same as that of Clothing1M.
\subsection{Clothing1M}

\begin{table}
\setlength{\tabcolsep}{1.3mm}
\small
\begin{center}
\begin{tabular}{c|c|c|c}
\hline
\# & Method & Data & Accuracy\\
\hline
\hline
1& Cross Entropy &1M noisy & 69.54 \\
2& Forward~\cite{patrini2017making} & 1M noisy & 69.84 \\
3& Joint Optim.~\cite{tanaka2018joint} & 1M noisy & 72.23 \\
4& MLNT-Teacher~\cite{li2019learning} & 1M noisy & 73.47 \\
5& Ours & 1M noisy & \bf{74.45} \\
\hline
6& Forward~\cite{patrini2017making} & 1M noisy  + 25k verify & 73.11 \\
7& CleanNet $w_{hard}$~\cite{lee2018cleannet} & 1M noisy + 25k verify & 74.15\\
8& CleanNet $w_{soft}$~\cite{lee2018cleannet} & 1M noisy + 25k verify & 74.69\\
9& Ours& 1M noisy + 25k verify & \bf{76.44}\\
\hline
10& Cross Entropy &1M noisy+ 50k clean & 80.27 \\
11& Forward~\cite{patrini2017making} & 1M noisy + 50k clean & 80.38\\
12&CleanNet $w_{soft}$~\cite{lee2018cleannet} & 1M noisy + 50k clean & 79.90 \\
13& Ours & 1M noisy + 50k clean & \bf{81.16} \\
\hline
\end{tabular}
\end{center}
\vspace{-2mm}
\caption{The classification accuracy (\%) on Clothing1M compare with other methods.}
\label{clothing-comp}
\vspace{-4mm}
\end{table}
\begin{figure*}[h]
    \centering
    \small
    \begin{tabular}{c@{\hspace{-4mm}}c@{\hspace{3mm}}c@{\hspace{2mm}}c}
        &
        \begin{tabular}{c}
            \includegraphics[width=5.2cm]{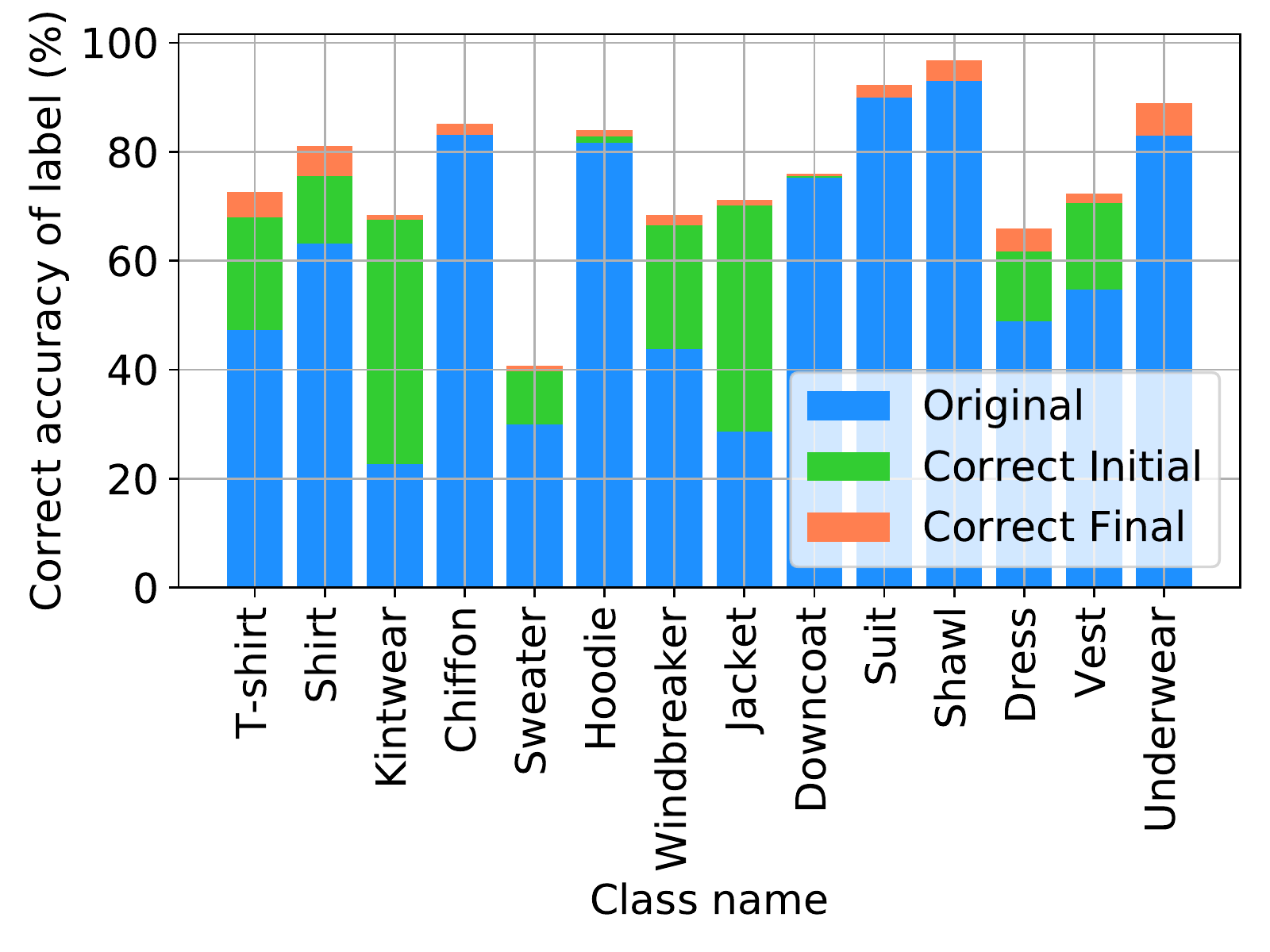}
        \end{tabular}
        &
        \begin{tabular}{c}
            \includegraphics[width=5.2cm]{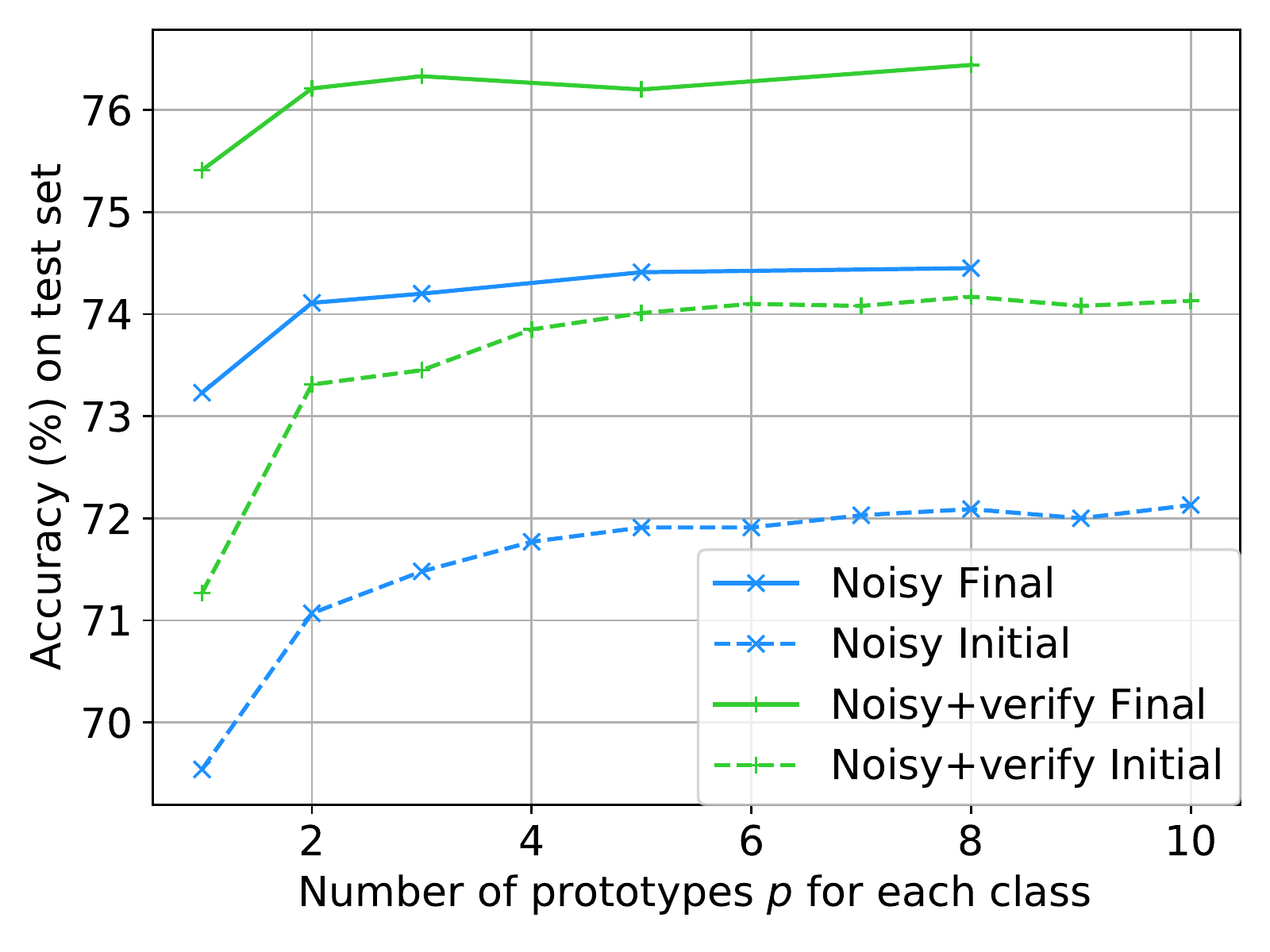}
        \end{tabular}
        &
        \begin{tabular}{c}
            \includegraphics[width=5.2cm]{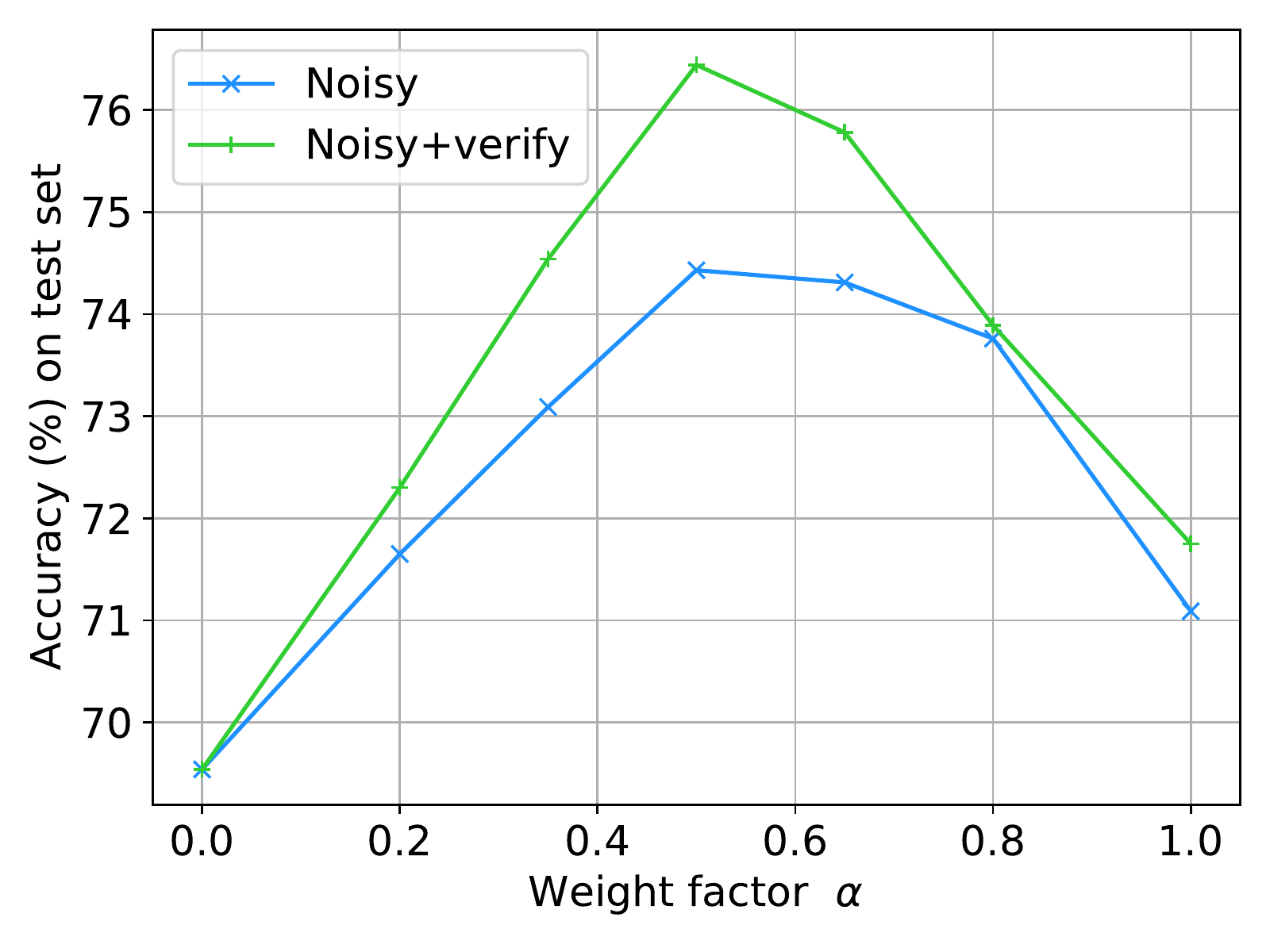}
        \end{tabular}\\
        &(a) & (b)& (c)\
    \end{tabular}
   \caption{(a) The label accuracy (\%) of labels in the original dataset (Original), labels corrected by the label correction phase in the first iterative cycle (Correct Initial) and labels corrected by the model at the end of training (Correct Final) for each class in Clothing1M. (b) Testing accuracy (\%) with the number of prototypes $p$ ranging from 1 to 10 for each class. The solid line denotes the accuracy got by the model at the end of training (Final). The dotted line denotes the correct accuracy by the model just step into the label correction phase for the first time (Initial). Noisy is the result of the training from noisy dataset only, noisy+verify indicates additional verification information is used. (c) Testing accuracy (\%) with weight factor $\alpha$ ranging from 0.0 to 1.0. Noisy and noisy+verify have the same meaning as (b).}
\label{fig:mix}
\vspace{-4mm}{}
\end{figure*}
We adopt the following three settings by following previous work.
First, only noisy dataset is used for training without using any extra clean supervision in the training process.
Second, verification labels are provided, but they are not used to train the network directly. \eg They are used to train the accessorial network as~\cite{lee2018cleannet} or to help select prototypes in our method.
Third, both noisy dataset and 50k clean labels are available for training.

We compare the results in Table~\ref{clothing-comp}.
We see that in the first case,
the proposed method outperforms the others by a large margin, \eg improving the accuracy from 69.54\% to 74.45\%, better than Joint Optimization~\cite{tanaka2018joint} (\#3) by 2.22\% and MLNT-Teacher~\cite{li2019learning} (\#4) by 0.98\%.
Our result is even better than \#6 and \#7 which uses extra verification labels.

For the second case, Traditional Cross-Entropy is not suitable. \cite{patrini2017making} used the information to estimate the transition matrix, while
CleanNet \cite{lee2018cleannet} used the verification labels to train an additional network to predict whether the label is noisy or not.
Our method uses this information to select the class prototypes.
In this case, we still achieve the best result compared to all methods.

For the third case, all data (both noisy and clean) can be used for training.
All the methods first train a model on the noisy dataset and then the model is finetuned using vanilla cross-entropy loss on the clean dataset.
We see that our method still outperforms the others. CurriculumNet~\cite{guo2018curriculumnet} provides a slightly better result (81.5\%) in this case. But it uses a different backbone compared with all others, so we do not consider it.
Among all of these cases, our approach obtains state-of-the-art performances compared to previous methods, showing that our method is effective and suitable for board situations.
\subsection{Ablation Study}

\textbf{Label Correction Accuracy.}
We explore the classification accuracy in the label-correction phase.
Table~\ref{clothing-noisy-rate} lists the overall accuracy in the original noisy set: the accuracy of the corrected label in the initial iterative cycle (\ie. the first time we step into the label correction phase after training the preliminary model), and accuracy of the corrected label by the final model at the end of training (Final).
We see that the accuracy after the initial cycle already reaches 74.38\%, improving the original accuracy by 12.64\% (61.74\% \vs 74.38\%).
The accuracy is further improved to 77.36\% at the end of the training.
%
%

\begin{figure}[t]
\begin{center}
\includegraphics[width=8cm ,scale=0.56]{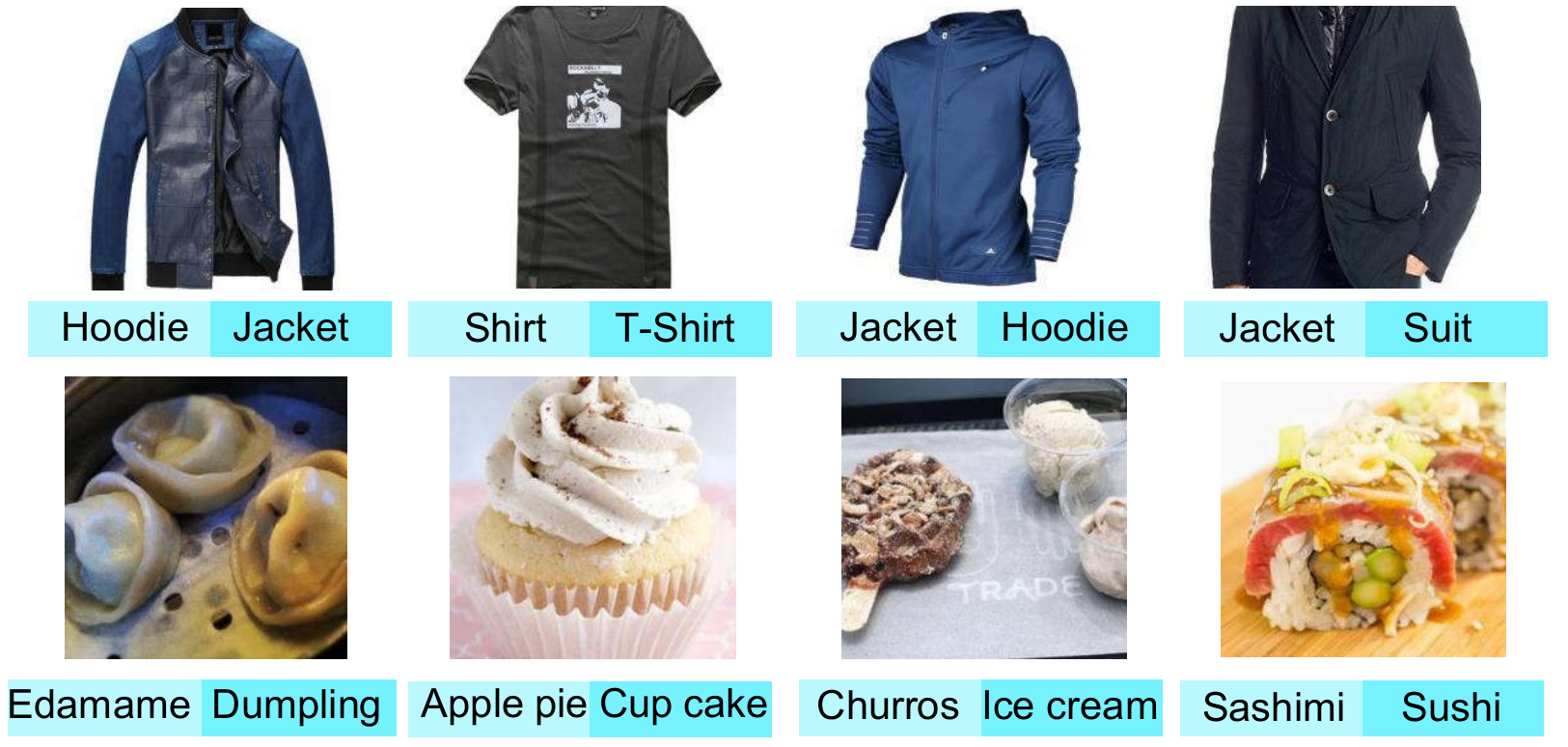}
\end{center}
\vspace{-4mm}
   \caption{Samples corrected by our method. \textbf{Left}: The original noisy label. \textbf{Right}: The right label corrected by our method. The first row from Clothing1M and the second row from Food101-N.}
\label{fig:vis_case}
\vspace{-5mm}{}
\end{figure}

We further explore the classification accuracy for different classes as shown in Figure~\ref{fig:mix}~(a).
We can find that for the most classes with the original accuracies lower than 50\%, our method can improve the accuracy to higher than 60\%.
Even for the 5th class (``Sweater'') with about 30\% original accuracy, our method still improves the accuracy by 10\%.
Some of the noisy samples successfully corrected by our approach are shown in Figure~\ref{fig:vis_case}.

\begin{table}
\setlength{\tabcolsep}{1.3mm}
\small
\begin{center}
\begin{tabular}{c|c|c|c}
\hline
 & Original & Correct Initial &Correct Final  \\
\hline
Accuracy &61.74 & 74.38 &\bf{77.36} \\
\hline
\end{tabular}
\end{center}
\vspace{-2mm}
\caption{Overall label accuracy (\%) of the labels in original noisy dataset (Original), accuracy of the corrected label generated by the label correction phase in first iterative cycle (Correct Initial) and accuracy of the corrected labels generated by the final model when training ends (Correct Final).}
\label{clothing-noisy-rate}
\vspace{-3mm}
\end{table}

\textbf{Number of Class Prototypes $p$}.
The number of class prototypes is the key to the representation ability to a class.
When $p=1$, the case is similar to CleanNet~\cite{lee2018cleannet}. In our method, we use $p \geq 1$. Another difference is that CleanNet attained the prototype by training an additional network. But we just need to select images as prototypes by their density and similarity according to the data distribution.

Figure~\ref{fig:mix}~(b) shows the effect of changing the number of prototypes for each class.
We select five $p$ values and evaluate the final test accuracy trained by either only using 1M noisy data or adding 25k verification information, as shown by the solid lines.
To have a better observation of the influence caused by $p$ value, we evaluate the label correction rate by the model step into the first label correction phase, which is similar to the correction accuracy discussed in the last experiment. 
But this time we evaluate on the testing set.
This metric is easy to be evaluated, so we explore 10 $p$ values from 1 to 10, as shown in the dotted line.
When comparing these two settings, they follow the same trend.

From the result, we find that when $p=1$ \ie one prototype for each class,
the accuracies are sub-optimal compared to others.
When using more prototypes, the performance improves a lot, \eg the accuracy using two prototypes outperforms using a single one by 2.04\%
This also proves our claim that a single prototype is not enough to represent the distribution of a class.
Multiple prototypes provide more comprehensive representation to the class.

\textbf{Weight factor $\alpha$}.
Weight factor $\alpha$ plays an important role in the training procedure, which decides the network will concentrate on the original noisy labels $Y$ or on the corrected labels $\hat{Y}$.
If $\alpha = 0$, the network is trained by using only noisy labels without correction.
Another extreme case is when $\alpha = 1$, the training procedure discards the original noisy labels and only depends on the corrected labels.
We study the influence of different $\alpha$ ranging from 0.0 to 1.0 and the test accuracy with different $\alpha$ is shown in Figure~\ref{fig:mix}~(c).

From the result, we find that training using only the noisy label $Y$ \ie $\alpha=0$ leads to poor performance.
Although the corrected label is more precise, the model trained using only corrected label $\hat{Y}$ also performs sub-optimal.
The model jointly trained by using the original noisy label $Y$ and the corrected label $\hat{Y}$ achieves the best performance when $\alpha=0.5$.
The accuracy curve also proves our claim that label correction may misrecognize some hard samples as noises.
Directly replacing all noisy labels with the corrected ones would make the network focused on simple features and thus degrade the generalization ability.

\textbf{Number of Samples $m$}.
To avoid massive calculation related to the similarity matrix $\mathbf{S}$,
we randomly select $m$ images rather than using all of the images in the same class to compute the similarity matrix.
We examine how many samples are enough to select the class prototypes to represent the class distribution well.
We explore the influence of the images number $m$ for each class.
\begin{table}
\setlength{\tabcolsep}{1.3mm}
\small
\begin{center}
\begin{tabular}{c|c|c|c|c}
\hline
$m$ & 320 & 640 & 1280& 2560 \\
\hline
\hline
1M noisy (Final) & 74.37 & 74.07 & \bf{74.45} & 74.27 \\
1M noisy (Initial)& 72.04 & 72.03 & \bf{72.09} & 72.05 \\
\hline
1M noisy + 25k verify (Final)& 76.43 & 76.49 & 76.44 &\bf{76.55} \\
1M noisy + 25k verify (Initial)& 74.09 & 73.97 & 74.17 &\bf{74.21} \\
\hline
\end{tabular}
\end{center}
\vspace{-2mm}
\caption{The classification accuracy (\%) on Clothing1M with different number of samples used to select prototypes for each class. Final denotes the accuracy get by the model at the end of training. Initial denotes the correct accuracy by the model just step into the first label correction phase.}
\vspace{-5mm}
\label{clothing-num}
\end{table}

The results are listed in Table~\ref{clothing-num}.
Experiment setting is similar to the experiments above to study the number of class prototypes.
The models are trained on the noisy dataset as well as the noisy dataset plus extra verification labels respectively.
The results are the accuracy of the trained model on the test set.
Besides evaluating the classification accuracy got by the final model, we also examine the correction accuracy of the model just step into the first label correction phase, which is denoted by ``Initial'' in the table.
By analyzing the results in different cases, we see that the performance is not sensitive to the number of images $m$.
Compared with 70k training images in Clothing1M for each class, we merely sample 2\% of them and obtain the class prototypes to represent the distribution of the class well.

\textbf{Prototype Selection}.
To explore the influence of the method used to select prototypes, we also use two other clustering methods to get the prototypes.
One is the density peak by Euclidean distance~\cite{rodriguez2014clustering}, while
the other is the widely used K-means algorithm, that is, K-means++~\cite{arthur2007k}.
The prototypes attained by all the methods are used to produce the corrected labels for training.
Results are listed in Table~\ref{clothing-cluster}.
\begin{table}
\setlength{\tabcolsep}{1.3mm}
\small
\begin{center}
\begin{tabular}{c|c|c}
\hline
Method & Data & Accuracy  \\
\hline
\hline
K-means++~\cite{arthur2007k} & 1M noisy&74.08\\
Density peak Euc.~\cite{rodriguez2014clustering}& 1M noisy &  74.11\\
Ours & 1M noisy & \bf{74.45}\\
\hline
K-means++~\cite{arthur2007k} & 1M noisy + 25k verify&76.22\\
Density peak Euc.~\cite{rodriguez2014clustering}& 1M noisy + 25k verify& 76.05 \\
Ours & 1M noisy + 25k verify & \bf{76.44}\\
\hline
\end{tabular}
\end{center}
\vspace{-2mm}
\caption{The classification accuracy (\%) on Clothing1M with different cluster methods used to select the prototypes.}
\label{clothing-cluster}
\end{table}
We see that the method used to generate prototypes does not largely impact the accuracy, implying that
our framework is not sensitive to the clustering method.
But the selection method proposed in this work still performs better than others.
\subsection{Food-101N}
%
\begin{table}
\setlength{\tabcolsep}{1.3mm}
\small
\begin{center}
\begin{tabular}{c|c|c}
\hline
\# & Method &  Accuracy\\
\hline
\hline
1& Cross Entropy & 84.51 \\
2& CleanNet $w_{hard}$~\cite{lee2018cleannet} &  83.47\\
3& CleanNet $w_{soft}$~\cite{lee2018cleannet} &  83.95\\
4& Ours& \bf{85.11}\\
\hline
\end{tabular}
\end{center}
\vspace{-2mm}
\caption{The classification accuracy (\%) on Food-101N compare with other methods.}
\vspace{-4mm}
\label{food-comp}
\end{table}
We also evaluate our method on the Food-101N~\cite{lee2018cleannet} dataset.
The results are shown in Table~\ref{food-comp}.
We find that our method also achieves state-of-the-art performance on Food-101N, outperforming CleanNet~\cite{lee2018cleannet} by 1.16\%.
\vspace{-1mm}
\section{Conclusion}
\vspace{-1mm}
In this paper, we propose an iterative self-learning framework for learning on the real noisy dataset.
We prove that a single prototype is insufficient to represent the distribution of a class and multi-prototypes are necessary.
We also verify our claim that original noisy labels are helpful in the training procedure although the corrected labels are more precise.
By correcting the label using several class prototypes and training the network jointly using the corrected and original noisy iteratively,
this work provides an effective end-to-end training framework without using an accessorial network or adding extra supervision on a real noisy dataset.
We evaluate the methods on different real noisy datasets and obtain state-of-the-art performance.
\vspace{-1mm}
\section*{Acknowledgement}
\vspace{-1mm}
This work is supported in part by SenseTime Group Limited, in part by the General Research Fund through the Research Grants Council of Hong Kong under Grants CUHK14202217, CUHK14203118, CUHK14205615, CUHK14207814, CUHK14213616.

{\small
\bibliographystyle{ieee_fullname}
\bibliography{egbib}
}

\end{document}